\documentclass[11pt]{article}

\usepackage{acl}

\usepackage{times}
\usepackage{latexsym}

\usepackage[T1]{fontenc}

\usepackage[utf8]{inputenc}

\usepackage{microtype}

\usepackage{inconsolata}

\usepackage{graphicx}

\usepackage{graphicx}
\usepackage{pgfplots}
\usepackage{times}
\usepackage{latexsym}
\usepackage[T1]{fontenc}
\usepackage[utf8]{inputenc}
\usepackage{enumitem}
\usepackage{microtype}
\usepackage{inconsolata}
\usepackage{graphicx}
\usepackage{csquotes}
\usepackage{xspace}
\usepackage{amssymb}
\usepackage{booktabs}
\usepackage{tabularx}
\usepackage{pgfplots}
\usepackage{url}
\usepackage{todonotes}
\usepackage{amsmath}
\usepackage{placeins} %
\usepackage{fix-cm}
\usepackage{subcaption}
\usepackage{multirow}
\usepackage{algorithm}
\usepackage{algpseudocode}
\usepackage{newunicodechar}
\usepackage{pifont}
\usepackage{booktabs,multirow,siunitx}
\newunicodechar{★}{\ding{72}}
\pgfplotsset{compat=1.18}
\usepackage{tikz}

   \everymath=\expandafter{\the\everymath\displaystyle}
   \IfFileExists{scrextend.sty}{
   \usepackage[fontsize=10.000000pt]{scrextend}
  }{
     \renewcommand{\normalsize}{\fontsize{10.000000}{12.000000}\selectfont}
     \normalsize
   }
   
   \makeatletter\@ifpackageloaded{underscore}{}{\usepackage[strings]{underscore}}\makeatother

\newcommand{\bracketcounter}[1]{\textbf{(#1)}}

\usepackage{tcolorbox}
\usepackage{xcolor}
\tcbuselibrary{skins,breakable}

\ifdefined\NavyBlue\else
  \definecolor{NavyBlue}{rgb}{0.0, 0.0, 0.5}
\fi

\ifdefined\chapter
  \newcounter{mdquote}[chapter]
  
\else
  \newcounter{mdquote}[section]
  
\fi

\newtcolorbox{mdquote}[1][]{
  enhanced,
  breakable,
  before upper={\refstepcounter{mdquote}},
  colback=blue!5,
  colframe=black!15,
  boxrule=0pt,
  left=12pt,
  right=12pt,
  top=6pt,
  bottom=6pt,
  borderline west={2pt}{0pt}{NavyBlue},
  #1
}

\title{The Canonical Order Problem: When Large Language Models Are Unreliable Knowledge Bases for Multi-Valued Relations}

\author{Timo Pierre Schrader$^{1}$~
  Annemarie Friedrich$^1$~
  Simon Razniewski$^{2}$~
  Lukas Lange$^3$\\
    $^1$University of Augsburg, Augsburg, Germany \\
    $^2$ScaDS.AI \& TU Dresden, Dresden, Germany \\
    $^3$Bosch Center for Artificial Intelligence, Renningen, Germany \\ 
\texttt{lukas.lange@de.bosch.com}}

\begin{document}
\maketitle
\begin{abstract}
Large language models (LLMs) are increasingly used as knowledge bases (KBs) due to the vast amount of knowledge they acquire during pre-training. While many works focus on extracting single relational triples, most real-world relations are multi-valued and require generating sets of entities.

In this paper, we investigate how LLMs represent and generate multi-valued relations. We identify the \textbf{canonical order problem}: The probabilistic distributions inside LLMs organize many multi-valued relations according to a canonical ordering (e.g., alphabetical or chronological). Through mechanistic analysis, we show that set generation in LLMs can be thought of in terms of three phases: \bracketcounter{1} retrieval of candidate entities, \bracketcounter{2} internal sorting, and \bracketcounter{3} selection of the next element. 
As a result, prompts aiming to construct KBs that deviate from this internal canonical ordering  lead to a markedly reduced reliability of LLMs when aiming to generate complete sets for multi-valued relations. %
\end{abstract}

\section{Introduction}
Due to their pre-training on huge amounts of text, large language models (LLMs) possess broad knowledge implicitly stored in their parameters.
This renders LLMs particularly interesting for the extraction of structured knowledge to construct knowledge bases \citep[KBs,][]{hu2024gptkb,pinto2025exploring}.
Most existing work focuses on single-valued relations \citep{petroni-etal-2019-language,jiang-etal-2020-know,singhania-etal-2023-extracting}.
In practice, much relational knowledge is organized into sets such as \enquote{all Formula 1 world champions,} \enquote{all official EU languages,} or \enquote{all Turing Award winners.}
To extract this knowledge from an LLM, the model has to generate a set of entities. %
In practice, this leads to two main issues: 
(1) elements are \textit{omitted} during generation (\textit{Recall problem}) and
(2) the generation of additional wrong elements (\textit{Precision problem}).

Despite the practical importance of multi-valued relations, little is known about how LLM parameters store such information, or what happens when we attempt to retrieve it via prompting and generation.
In this paper, we study the internal mechanisms and challenges when extracting this type of relational knowledge from LLMs. 
Our findings reveal previously overlooked limitations of using LLMs for KB construction: whether or not they succeed in retrieving multi-valued relations depends strongly on the examples provided in the prompt. %

\begin{figure}[t]
    \centering
    \resizebox{1.0\linewidth}{!}{\input{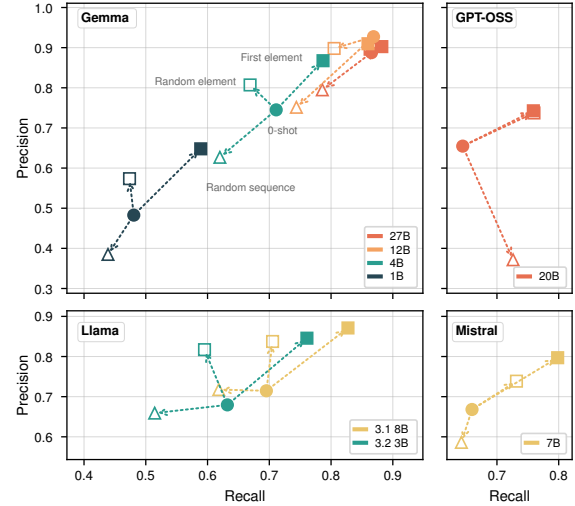}}
    \caption{
    Precision-recall curves for LLM set generation across 37 topics. 
    We compare four distinct settings: 
    \ensuremath{\bullet} without any %
    examples;
    \ensuremath{\blacksquare} with the first element in canonical order;
    \ensuremath{\square} with a single random element;
    and \ensuremath{\triangle} with randomly shuffled elements.
    Providing the first element as an example always achieves the best results.
    }
    \label{fig:teaser}
\end{figure}

We identify the \textbf{canonical order problem}: LLMs internally organize multi-valued relations according to a canonical order (e.g., numerical, alphabetical, or chronological), which causes poor reliability when the generation process deviates from this order.
This finding shows that in this context, LLMs do not \enquote{store} relational information consistently but rather that their output depends on whether prompts reflect canonical orders observed during pretraining.
As an example, we consider retrieving all 16 states of Germany.
If we ask a model such as Gemma-3 12B to come up with all 16 states, it will most likely generate all of them in alphabetical order.
However, if we ask it to complete a random subset of 9 states, it for example tends to forget \textit{Bremen} in its final output.
Figure~\ref{fig:teaser} illustrates this effect across 37 topics and multiple model families.
We compare four prompting strategies: no example at all, the first element in canonical order, a single randomly chosen element, and a set of randomly shuffled elements.
In all cases, providing the first element in canonical order yields the most complete and accurate sets, while forcing the model to continue from shuffled elements leads to substantially more omissions and errors.

To better understand this model behavior, we perform a mechanistic analysis.
We analyze the autoregressive generation process by inspecting the outputs of each Transformer block and how expected entities behave in the hidden states.
Our mechanistic analysis suggests that activation in earlier Transformer layers reflect all set entities equally (reflecting the overall topic similar to a \textbf{retrieval step}) and that the following layers increasingly reflect canonical orderings (\textbf{sorting}), which becomes problematic when prompts propose example set entities that are not at the beginning of the canonical ordering. %
This implies that multi-valued %
knowledge extraction from LLMs is more reliable when they do not have to deviate from a canonical order.
We argue that the observed behavior limits the reliability of zero-shot LLMs as KBs, since the correctness of generated sets depends not only on the stored facts but also on the generation order.

\section{Related Work}
In this section, we review related work on using LLMs for KB construction, using LLMs in combination with structured knowledge and multi-valued relations, ordering problems in LLMs, and approaches to mechanistic analyses of LLMs.

\paragraph{Language Models for KB Construction}
The knowledge implicitly stored in language models is widely used for constructing KBs \citep{petroni-etal-2019-language,alkhamissi2022reviewlanguagemodelsknowledge}.
GPTKB \citep{hu2024gptkb} extracts 105 million relations from GPT-4o mini, greatly exceeding the number of entries in well-known KBs such as Wikidata \citep{wikidata} and YAGO \citep{yago}.
A multi-turn scheme for extracting relational triples is introduced by \citet{zhang2024extract}, where free-form LLM-based extraction is followed by canonicalization of generated relations into a general schema.

Another branch of KB construction uses the language understanding capabilities of LLMs to extract relational triples from documents.
A pipeline for extracting relational triples from text given a priori retrieved context from an existing KB is proposed by \citet{papaluca-etal-2024-zero}.
Applications in materials science demonstrate how LLMs can extract knowledge from tabular data for KG construction \cite{dreger2025large}.
Earlier approaches leverage encoder-based language models to treat relation extraction as a classification task \cite{wei-etal-2020-novel}.

We position our work as an analysis of the intermediate step of knowledge elicitation from LLMs before actually building up a KB.
Although the generated sets of tail entities could be used for constructing a KB, we aim to understand how LLMs internally represent and generate relational knowledge.

\paragraph{LLMs and Multi-Valued Relations}
There are two main facets when dealing with multi-valued relations in the context of LLMs: \textit{extraction} and \textit{editing}.
On the extraction side, \citet{singhania-etal-2023-extracting} probe BERT encoders \citep{devlin-etal-2019-bert} using masked language modeling to generate candidate entities.
Different heuristics for selecting candidate sets are explored, such as top-$k$ selection or probability thresholds derived from model logits.
In contrast, our work focuses on analyzing set generation in autoregressive decoder-based models rather than proposing selection heuristics.
Decoder-based LLMs benefit from broader general-purpose knowledge due to next-token prediction objectives and typically larger parameter counts.
On the editing side, encoder-decoder models are fine-tuned to generate either individual instances or complete sets \cite{nagasawa-etal-2023-lms}.
While high output accuracy can be achieved, this often comes at the cost of reduced generalizability.

\paragraph{LLMs and Structured Knowledge}
Research in this area investigates how structured knowledge can be processed, generated, or integrated with LLMs.
This includes tabular data \cite{agarwal-etal-2025-hybrid,zhou-etal-2025-efficient,liu-etal-2024-rethinking}, KGs \cite{zhou-etal-2025-reflection,regino-dos-reis-2025-llms,zhu-etal-2025-knowledge}, databases \cite{qin2024relationaldatabaseaugmentedlarge,li2023can}, natural language documents via retrieval-augmented generation \cite{10887211,arslan2024survey}, and ontologies \cite{mai2024llms,10.1145/3696410.3714816}.
Our work focuses on structured knowledge in the form of lists, which may be stored in documents or generated directly as responses.

\paragraph{Influence of Orderings in Language Models}
The ordering of few-shot demonstrations has been shown to strongly affect the accuracy of autoregressive language models, for example, the ordering in which demonstrations are presented to models in few-shot prompts \cite{lu-etal-2022-fantastically,zhao2021calibrateuseimprovingfewshot}.
It has also been shown that learning relational knowledge of the form $A \equiv B$ does not necessarily imply learning the reverse relation $B \equiv A$ \cite{berglundreversal}.
The effect of input ordering on reasoning performance is further explored in multi-document question answering and logical reasoning tasks \cite{liu-etal-2024-lost,pmlr-v235-chen24i}.
Performance degrades when relevant information appears later in the prompt but improves again when placed near the end, forming a U-shaped curve \cite{liu-etal-2024-lost}.
Similarly, deviations from natural premise ordering negatively impact reasoning accuracy, especially as the number of premises increases \cite{pmlr-v235-chen24i}.
While existing work focuses on input ordering, we investigate the extent to which ordering is inherent in stored knowledge.

\paragraph{Mechanistic Analysis of LLMs}
Understanding the internal mechanisms of language models has been a long-standing goal in interpretability research.
A major line of work focuses on attribution methods that assign importance scores to input features.
Integrated Gradients \cite{DBLP:conf/icml/SundararajanTY17} is a widely used method based on axiomatic principles such as sensitivity and implementation invariance.
Second-order methods have also been proposed to analyze interactions between modalities in dual encoder models \cite{moeller2025explaining}, such as CLIP \cite{radford2021learningtransferablevisualmodels}.

In language models, attention analysis is commonly used to study information flow in Transformer architectures \cite{abnar-zuidema-2020-quantifying,clark-etal-2019-bert,jain-wallace-2019-attention}.
In this work, we employ the logit lens technique, which projects hidden states onto the vocabulary space by applying the final normalization and language modeling head.
This method has been used to analyze where factual knowledge is stored in multilingual LLMs \cite{wang-etal-2025-lost-multilinguality}, to study social biases \cite{prakash-lee-2023-layered}, and to investigate visual feature representations in vision-language models \cite{huo-etal-2024-mmneuron}.

\section{Theoretical Concepts}
\label{app:theoretical_concepts}
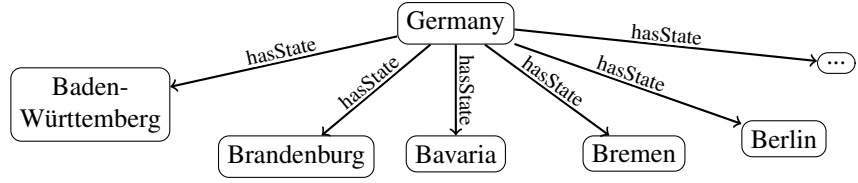
\begin{figure*}[!ht]
    \centering

\begin{tikzpicture}[
    node distance=1.4cm,
    every node/.style={rounded corners, align=center},
    arrow/.style={->, thick},
    edgelabel/.style={midway, sloped, fill=white, inner sep=1pt, font=\footnotesize}
]

\node[draw] (root) {Germany};

\node[draw, below left=3mm and 3cm of root] (bw) {Baden-\\Württemberg};
\node[draw, below=1.2cm of root] (by) {Bavaria};
\node[draw, below right=10mm and 3cm of root] (be) {Berlin};
\node[draw, below left=12mm and 0.3cm of root] (bb) {Brandenburg};
\node[draw, below right=12mm and 0.9cm of root] (hb) {Bremen};

\node[draw, below right=1mm and 4cm of root] (etc) {...};

\draw[arrow] (root) -- node[edgelabel, above] {hasState} (bw);
\draw[arrow] (root) -- node[edgelabel, above] {hasState} (by);
\draw[arrow] (root) -- node[edgelabel, above] {hasState} (be);
\draw[arrow] (root) -- node[edgelabel, above] {hasState} (bb);
\draw[arrow] (root) -- node[edgelabel, above] {hasState} (hb);

\draw[arrow] (root) -- node[edgelabel, above] {hasState} (etc);

\end{tikzpicture}

    \caption{Example demonstrating a multi-valued relation in a KB. The dependent nodes share the same head node and are connected to it via the same relation type.}
    \label{fig:multi-valued relation}
\end{figure*}

\paragraph{Background}
We define multi-valued relations %
as sets of entities $\mathcal{S}(t) = \{e_1, \dots, e_n \}$ that are in an \enquote{isA} relation with head node $h$.
Such relations can be represented as sparse or dense KGs with one node for each entity connecting it to the head as illustrated in Figure \ref{fig:multi-valued relation}.

Following standard relation extraction terminology, we also call the head node the relations \textit{subject} and the dependent node its \textit{entities}.

\paragraph{Subsets and Prefixes}
We define \textit{set prefixes} to be an unordered true subset $\mathcal{P}(t') \subset \mathcal{S}(t)$.
Although we use the term \enquote{prefix}, these subsets carry no ordering information; the term reflects their role as a starting point that the model must continue during generation.
For example, a prefix for the states in Germany might be $\mathcal{P}(t') = \{\text{Bremen}, \text{Saxony}\}$.
Set prefixes will be used in this paper to test how robustly LLMs can complete true subsets of the expected set while deviating from a canonical order.
Starting with such prefixes reflects a realistic scenario in knowledge base curation: practitioners typically have partial knowledge about the multi-valued relations they aim to collect.
For example, a knowledge engineer might already know that Bavaria %
is a state in Germany, and this partial information naturally suggests exploring related queries, such as \enquote{Besides Bavaria, what other states are located in Germany?}
This approach is motivated by few-shot learning methods where providing explanatory examples significantly improves model performance and guides the completion process~\cite{NEURIPS2020_1457c0d6,jiang2025fstllm}.
By starting with $k$ prefix elements, we simulate this workflow while testing how robustly LLMs can complete true subsets despite deviating from canonical orderings.
In our experiments, set prefixes enable us to analyze \bracketcounter{1} how well LLMs can extend partial knowledge, and \bracketcounter{2} whether the presence of non-canonical orderings in the prefix disrupts the completion of remaining set elements.

\paragraph{Auto-Regressive LLM Generation Process}
LLMs are statistical distributions that model the probability of the next token at each decoding step $j$:
\begin{align*}
    LLM: V \mapsto [0,1], \qquad \sum_{t \in V} \mathbb{P}(T_j = t) = 1
\end{align*}

This mapping assumes a normalization function that is applied on the LLM logits such as softmax. %
The joint distribution over a sequence of tokens $(T_1,\dots,T_n)$ can be decomposed using conditional probabilities:
\begin{align*}
    \mathbb{P}(T_1,\dots,T_n)
    = \mathbb{P}(T_1)\,\prod_{j=2}^n \mathbb{P}(T_j\mid T_1,\dots,T_{j-1})
\end{align*}

This decomposition allows us to inspect each auto-regressive decoding step $j$ individually.
Hence, we can inspect how likely each token $t \in V$ is to be generated at each step.
We inspect these probabilities to learn where \enquote{missed} elements that are never accurately verbalized are ranked in the probability distribution during generation.
Furthermore, due to the discrete generation process, we can determine \enquote{points of interest} at which we look deeper into the hidden states of a Transformer-based LLM, which is defined as the output of one Transformer block (attention combined with two linear layers).
We provide details in Section~\ref{sec:mechanistic_analysis_results}.

\section{Methodology}
\label{sec:mechanistic_analysis_methodology}
This section describes the creation of our dataset for prompting LLMs to extract sets of entities for given multi-valued relations.
We then analyze the LLMs' probability distribution while generating particular set items and analyze the relationship between missed items and canonical orderings. %

\paragraph{Dataset}
We first create a dataset that consists of 37 distinct multi-valued relations for which we can prompt an LLM to come up with all entities that occur as entities.
These relations, alongside the canonical orderings of the entities, are listed in Table~\ref{tab:dataset_topics}.
We ensure that each relation has a well-defined number of entities without frequent changes over time. %
We leave the analysis of non-stable sets for future work.
We sort the topics according to the size of the corresponding set of entities $\mid \mathcal{S}(t)\mid $ and categorize them into \textit{small ($\mid \mathcal{S}(t)\mid <15$)}, \textit{regular ($15\leq \mid \mathcal{S}(t)\mid  \leq 50$)}, and \textit{large ($\mid \mathcal{S}(t)\mid >50$)}.
Furthermore, we assign a \textbf{canonical ordering} to each $\mathcal{S}(t)$.
In this context, we refer to a canonical ordering as a commonly used way of ordering a list of entities, such as alphabetically, chronologically for temporal events, or according to other domain-specific criteria (e.g., the distance of planets from their sun).
We assign the canonical ordering based on the most common convention found in common literature or web sources (e.g., Wikipedia).

\begin{table*}
    \centering
    \setlength{\tabcolsep}{3pt}
    \renewcommand{\arraystretch}{1.35} %
    \scriptsize

    \vspace*{\fill}

    \begin{minipage}[t]{0.485\textwidth}
        \vspace{0pt}%
        \centering
        \begin{tabularx}{\linewidth}{@{}X c l l@{}}
            \hline
            \textbf{Topic} & \textbf{Count $\downarrow$} & \textbf{Ordering} & \textbf{Size} \\
            \hline
            Nobel prize categories & 6 & other & small \\
            Ancient wonders & 7 & chronological & small \\
            Newton spectrum colors & 7 & numerical & small \\
            Crystal systems & 7 & other & small \\
            Seven seas & 7 & other & small \\
            Moon phases & 8 & chronological & small \\
            Solar system planets & 8 & numerical & small \\
            Beethoven symphonies & 9 & chronological & small \\
            Poker hands & 10 & numerical & small \\
            German chancellors & 10 & chronological & small \\
            Zodiac signs & 12 & chronological & small \\
            Countries in SA & 13 & alphabetical & small \\
            Eclipse versions & 13 & chronological & small \\
            German states & 16 & alphabetical & regular \\
            MBTI 4-letter types & 16 & other & regular \\
            Fib. numbers ($\leq 1000$) & 17 & numerical & regular \\
            Standard amino acids & 20 & alphabetical & regular \\
            MS Office versions & 22 & chronological & regular \\
            India: official languages & 23 & alphabetical & regular \\
            \hline
        \end{tabularx}
    \end{minipage}\hfill
    \begin{minipage}[t]{0.485\textwidth}
        \vspace{0pt}%
        \centering
        \begin{tabularx}{\linewidth}{@{}X c l l@{}}
            \hline
            \textbf{Topic} & \textbf{Count $\downarrow$} & \textbf{Ordering} & \textbf{Size} \\
            \hline
            EU official languages & 24 & alphabetical & regular \\
            Bond films & 25 & chronological & regular \\
            Primes ($\leq 100$) & 25 & numerical & regular \\
            Mammal orders & 27 & numerical & regular \\
            NBA teams & 30 & alphabetical & regular \\
            F1 world champions & 34 & chronological & regular \\
            Shakespeare First Folio & 36 & chronological & regular \\
            NVIDIA RTX GPUs & 36 & chronological & regular \\
            U.S. presidents & 45 & chronological & regular \\
            HTTP status codes & 63 & numerical & large \\
            U.S. national parks & 63 & alphabetical & large \\
            Bible books & 66 & chronological & large \\
            Constellations & 88 & alphabetical & large \\
            Periodic table elements & 118 & numerical & large \\
            Primes ($\leq 1000$) & 168 & numerical & large \\
            Physics Nobel laureates & 230 & chronological & large \\
            Stable nuclides & 251 & numerical & large \\
            Catholic popes & 267 & chronological & large \\
            ~ & ~ & ~ & ~ \\
            \hline
        \end{tabularx}
    \end{minipage}

    \vspace*{\fill}

    \caption{The 37 distinct topics in our dataset along with their number of entities $\mid \mathcal{S}(t)\mid $, type of canonical ordering, and categorization into the three set sizes.}
    \label{tab:dataset_topics}
\end{table*}

\paragraph{Evaluation Framework}
When analyzing how reliably LLMs can generate complete sets of entities for multi-valued relations, we treat them as \textbf{unordered} sets and use precision (P), recall (R), and F1 score for evaluation.
To understand what happens inside an LLM during the autoregressive text generation process, we look at two distinct parts of the text generation process: \bracketcounter{1} a \textbf{single forward pass} including all hidden states produced by the Transformer blocks, 
and \bracketcounter{2} the evolution \textbf{across all autoregressive forward passes}, i.e., all steps $j$ at which the probability $\mathbb{P}(T_j = t \mid T_{j-1}\dots T_1)$ is computed for all tokens $t \in V$ in the vocabulary.
To perform a mechanistic analysis, we use the \textit{logit lens}\footnote{\url{https://www.lesswrong.com/posts/AcKRB8wDpdaN6v6ru/interpreting-gpt-the-logit-lens}} for interpreting the outputs of hidden states in LLMs.
For each hidden layer, the method takes the $k$-dimensional output vector, applies the model’s final normalization layer followed by the language-modeling head (a linear layer) that produces the $\mid V\mid $ logits for all tokens in the vocabulary, thus allowing for a language-level interpretation of the hidden states.
We can then search for all expected entities in the (sorted) logits by retrieving the position of the first subtoken of every entity.\footnote{Some entities may share the same initial subtoken, which can lead to overlapping matches. This is a limitation of the tokenizer design and introduces ambiguities when mapping entities to vocabulary tokens. Furthermore, we search for subtokens with and without leading whitespace and select whichever is higher ranked.}
When performing this search across sorted logits, we obtain a relative ranking $\hat{R}$ as well as an \textbf{absolute} rank between $0$ and $\mid V\mid -1$ for each entity.
We compare $R$, the ranking induced by the canonical ordering (i.e., sorting the entities by a fixed convention such as alphabetical or chronological), with $\hat{R}$, the ranking implied by the order in which the LLM generates the entities, using Spearman’s rank correlation coefficient $\rho$ to quantify agreement between the two orderings.
For the mechanistic analysis, we treat these sets as \textbf{ordered}.

\paragraph{Prompt Templates}
The base instruction for the model for every set generation experiment and the simulated start of the model answer are as follows:

\begin{mdquote}
\textbf{User:}\newline
\textit{Give me a list of all \texttt{<SUBJECT_NODE>}. The list output should be in bullet point format with the dash sign ('-') used as bullet point style.}
\newline \newline
\textbf{Assistant:}\newline
\textit{The following list contains all entities that belong to \texttt{<SUBJECT_NODE>}: \textbackslash n \newline
- \texttt{<OPTIONAL_PREFIX>}}
\end{mdquote}

For $k$-prefix experiments, we replace \texttt{<OPTIONAL\_PREFIX>} with the bullet point list that holds all $k$ prefix elements.

\paragraph{String Comparison}
Finding matching strings in free-text generation is difficult since exact matches are not guaranteed due to the model having the freedom of casing, adding punctuation marks and markdown markers, spacing, etc.

In order to make the set comparison between $\mathcal{S}(t)$ and $\mathcal{\hat{S}}(t)$ as robust as possible, we use the \texttt{difflib.SequenceMatcher} implementation from Python, which builds upon the algorithm originally presented in \cite{ratcliff1988pattern}.
Our threshold is set to a strict value of $0.95$.
During our experimentation phase, this value has been shown to be a good trade-off.
We additionally perform some pre-processing steps that remove some special signs such as quotation marks and parentheses that do not contribute to any entity in a meaningful way.
The procedure is shown in Algorithm~\ref{alg:string_preprocessing}.

\begin{algorithm}
\caption{String Preprocessing}
\label{alg:string_preprocessing}
\begin{algorithmic}
\Require Input string $S_i$
\Ensure Normalized output string $S_o$
\State $S_o \gets strip\_whitespace(S_i)$
\State $S_o \gets extract\_from\_within\_md\_indicators(S_o)$
\State $S_o \gets remove\_enumeration\_signs(S_o)$
\For{each $delimiter$ in ["(", ":", " - ", " – ", " — ", "\textbackslash n"]}
    \State $S_o \gets S_o.split(delimiter)[0]$
\EndFor
\State $S_o \gets S_o.lower()$
\State $S_o \gets remove\_quotation\_marks(S_o)$
\State $S_o \gets squeeze\_multiple\_spaces\_into\_one(S_o)$
\end{algorithmic}
\end{algorithm}

\section{Results}
\label{sec:mechanistic_analysis_results}

We start by conducting multi-valued relation generation experiments to understand how canonical orderings influence the robustness.
We then provide mechanistic insights into the models to better understand why canonical orderings play a crucial role inside the LLMs.
For set generation experiments in Section \ref{ssec:set_generation}, we mainly focus on models from the Gemma-3 model family \cite{gemmateam2025gemma3technicalreport}.
We confirm the generalizability of our findings in an aggregated statistic for models from the Llama family \cite{grattafiori2024llama3herdmodels}, Mistral-0.3 7B \cite{jiang2023mistral7b}, and GPT-OSS 20B \cite{openai2025gptoss120bgptoss20bmodel}.
For the mechanistic analyses in Sections \ref{ssec:mech1} and \ref{ssec:mech2}, we use the instruction-tuned Gemma-3 12B model.

\subsection{Robustness of Set Generation}
\label{ssec:set_generation}
\textbf{We start with the question of how the choice of start element affects the reliability of set generation.} 
This provides a first indication of how LLMs are influenced by orderings.

To do that, we compare four settings: \bracketcounter{1} no start element, \bracketcounter{2} a start element from the first position, \bracketcounter{3} a single start element from any position, and \bracketcounter{4} an average across 25 runs with randomly shuffled start elements.
We embed these start elements by pre-filling them in the model's response so that the model is asked to generate all remaining entities while being forced to work with a prefix of $1$ or $k$ start elements.
Contrasting settings \bracketcounter{3} and \bracketcounter{4} isolates the effect of the \textit{amount} of deviation from the canonical order: a single misplaced element already perturbs generation, and the disruption grows as more shuffled elements are forced into the prefix.

\begin{table}[t]
\centering
\footnotesize
\sisetup{detect-weight=true}
\setlength{\tabcolsep}{2.5pt}
\resizebox{\columnwidth}{!}{%
\begin{tabular}{@{}l*{12}{S[table-format=2.1]}@{}}
\toprule
& \multicolumn{3}{c}{\textbf{1B}} & \multicolumn{3}{c}{\textbf{4B}} & \multicolumn{3}{c}{\textbf{12B}} & \multicolumn{3}{c}{\textbf{27B}} \\
\cmidrule(lr){2-4}\cmidrule(lr){5-7}\cmidrule(lr){8-10}\cmidrule(lr){11-13}
\textbf{Example(s)} & {S} & {R} & {L} & {S} & {R} & {L} & {S} & {R} & {L} & {S} & {R} & {L} \\
\midrule
0-shot
& 56.9 & 36.2 & 36.3
& 78.7 & 79.1 & 50.2
& \textbf{92.5} & \textbf{89.7} & 80.9
& 86.4 & 88.4 & \textbf{85.1} \\

First
& \textbf{61.3} & \textbf{51.8} & \textbf{36.4}
& \textbf{84.7} & \textbf{85.5} & \textbf{66.2}
& 92.2 & 87.4 & \textbf{82.6}
& \textbf{89.7} & \textbf{89.9} & 83.4 \\

1 Random
& 53.0 & 46.8 & 24.6
& 77.5 & 74.7 & 49.6
& 90.1 & 83.3 & 72.8
& 88.6 & 87.7 & 81.6 \\

$k$ Random
& 42.2 & 34.7 & 13.0
& 69.1 & 58.3 & 34.6
& 84.7 & 73.9 & 48.2
& 83.9 & 77.9 & 57.5 \\
\bottomrule
\end{tabular}
}
\caption{Average F1 (\%) for models in the Gemma-3 family, split by set sizes and prefix strategy. S = small, R = regular, L = large.}
\label{tab:f1_by_size_and_setting}
\end{table}

Table \ref{tab:f1_by_size_and_setting} shows the influence of different types of start elements across model sizes in the Gemma-3 family and across set sizes categories.
From the results, we observe that providing the first element according to a canonical ordering is beneficial compared to having no start element at all as it provides a strong signal about the expected ordering and composition of the multi-valued relation.\footnote{In the case of 0-prefixes, the LLMs often tend to be more verbose in their output, e.g., adding additional information in parentheses or writing full sentences after the bullet points.}
Furthermore, forcing the model to continue from randomly shuffled multiple start elements substantially decreases the robustness of set generation by a large margin, indicating that \enquote{disrupting} the canonical ordering leads to instabilities in the generation.
Figure \ref{fig:teaser} displays the same tendencies across more model families aggregated over all list size categories.

We further observe that the negative effect of shuffled prefixes grows with the size of the target set: the gap between the canonical (first-element) and shuffled settings is largest for large sets across all model sizes.
We hypothesize that this occurs because larger sets leave more room for the model to deviate from the canonical order, and the autoregressive pre-training can bias LLMs toward storing multi-valued relations as ordered sequences rather than independent triples.

\textbf{In sum, providing a single start element following the canonical ordering can elicit multi-valued relations from LLMs more robustly, while deviating from this order hurts the robustness.} %
In the following, we analyze mechanistically why sticking to the canonical order leads to better results.

\begin{figure}[!t]
    \centering
    \captionsetup{skip=2pt}

    \resizebox{\linewidth}{!}{\input{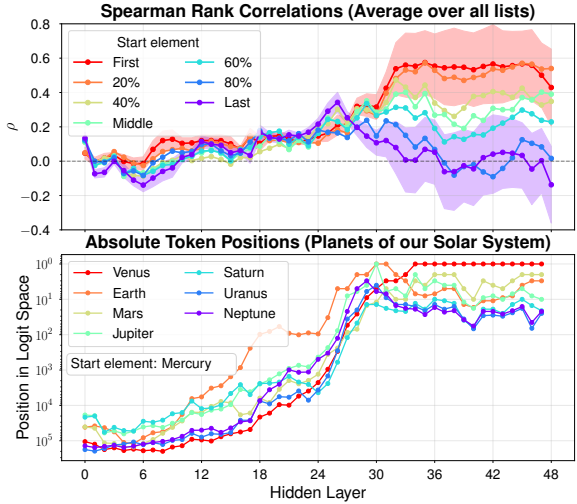}}
    
    \caption{
    \textbf{(a)} Evolution of the Spearman correlation between  expected and generated positions in each hidden layer of the first forward pass. 
    We evaluate different choices for the 1-prefix position.
    Shaded areas mark $\pm$ 1 standard deviation. 
    \textbf{(b)} Evolution of the absolute ranks in the sorted logit space for the 1-prefix \textit{Mercury}. Both 
    demonstrate that the canonical ordering emerges in the middle layers of the model.}
    \label{fig:logit_lens_first_decoding_step}
    \vspace{-0.6em}
\end{figure}

\subsection{Analysis of Prevalence of Canonical Orderings}
\label{ssec:mech1}
\textbf{In this section, we address the question of in which layers the canonical ordering of the target set is visible in the hidden states, and how this ordering behaves across consecutive forward passes.}
\paragraph{Single Forward Pass}

We begin by investigating the initial forward pass.
We instruct the model to generate the set of entities $\mathcal{\hat{S}}(t)$ for a given topic.
Furthermore, we extend the prompt with a short introductory phrase followed by a bullet point, ensuring that the first generated token corresponds to an entity in the set.

The top plot of Figure~\ref{fig:logit_lens_first_decoding_step} shows the evolution of the correlation coefficients $\rho$ in all $48$ hidden layers of Gemma-3-12B.
As explained in Section~\ref{sec:mechanistic_analysis_methodology}, we compute the correlation between the expected canonical order of entities and their actual relative ranks.
These values are averaged across the 12 topics in our dataset for which computation of all logit lens values is tractable.\footnote{ancient world wonders, books of the Bible, Nobel prize categories, countries in South America, distinct F1 world champions, MBTI personality types, official languages in India, official languages in the EU, phases of the moon, planets of our solar system, poker hands, states in Germany}

It becomes evident that up to approximately layer 25-30, there is almost the same correlation between the expected (inferred by the canonical order) and actual relative positions regardless of the input element.
Beginning from layer 30, the model begins to be influenced by which entity is provided as the starting example.
When using an element from the first position %
there is a medium-high correlation of around $0.5$, consistent with the model internally retrieving an ordered representation of the target set.
This correlation gradually drops the more the starting element is taken from the end of the set.
Interestingly, for a start element from the last position (e.g., \textit{Neptune}), there is no highly negative correlation. %
The visualized standard deviations of the red and purple lines further show that this finding holds across a variety of topics.
We interpret this as evidence that the model internally stores a representation aligned with a canonical ordering, while forcing it to deviate from it shows signs of the model losing any structured order. 
We hypothesize that this behavior arises primarily from the pre-training data distribution, where entities belonging to such sets are most often encountered in their canonical ordering.

\paragraph{Exemplary Topic}
To further understand the inner workings of set generation during the first forward pass, we examine the multi-valued relation \enquote{planets of our solar system} as an example using \textit{Mercury} as the starting element,  following the canonical ordering.
The bottom plot of Figure \ref{fig:logit_lens_first_decoding_step} depicts the absolute ranks of the logits corresponding to the remaining seven planets. Beginning around layer 20, the model starts to \enquote{pull} all necessary entities to top positions in the sorted logit space.

Referring to the bottom plot of Figure \ref{fig:logit_lens_first_decoding_step}, it becomes clear that once entities have been pulled to the top of the logit space, the model begins to sort them according to its internal ordering.
Summarizing the findings of both plots, we can define the following phases of multi-valued relation generation in LLMs: \textbf{(1) retrieval} of relevant entities, \textbf{(2) sorting} of the retrieved elements, and \textbf{(3) selection} of the next element to be generated.

\begin{figure}[t]
    \centering
    \resizebox{1.0\linewidth}{!}{\input{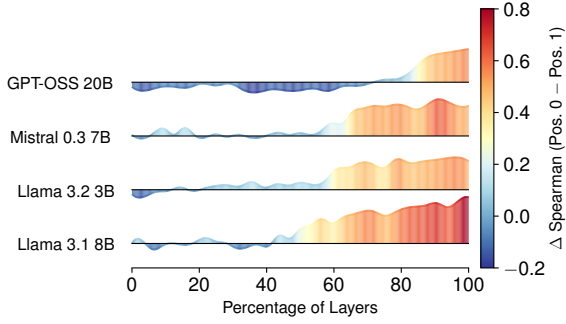}}
    \caption{Difference of correlation values $\rho$ when using the first and last element as start element across the hidden layers of the first forward pass.}
    \label{fig:spearman_ridgeline}
\end{figure}

\paragraph{Spearman Correlation Values for different Models}
Figure \ref{fig:spearman_ridgeline} shows the difference across hidden layers of the first forward pass between the correlation values $\rho$ when using the first element and the last element as starting element of the set.
We can see that our findings generalize to a variety of multi-valued relations and that approx. the last third of layers is where the internalized canonical order plays the most important role.

\begin{figure}[!t]
    \centering
    \resizebox{1.0\columnwidth}{!}{\input{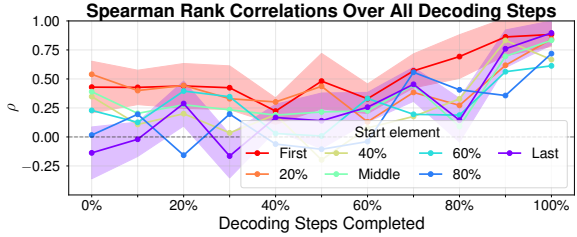}}
    \caption{Evolution of the Spearman correlation values between expected positions and actual positions across the last hidden states of all decoding steps. Shaded areas indicate $\pm$1 standard deviation.}
    \label{fig:logit_lens_across_decoding_steps}
\end{figure}

\paragraph{Across Forward Passes}
\label{ssec:across_decoding_steps}
Figure \ref{fig:logit_lens_across_decoding_steps} displays the average values of $\rho$ for the different starting elements described above but this time across the last hidden states of all decoding steps, i.e., the hidden states that are used to compute the model’s output predictions.
We observe very similar tendencies, with later start elements leading to low correlation with the canonical ordering.
Note that with an increasing number of generated set elements, the correlation with the canonical order for the remaining elements increases as there are fewer degrees of freedom for the model.

\begin{figure*}[!t]
    \centering
    \resizebox{0.8\textwidth}{!}{\input{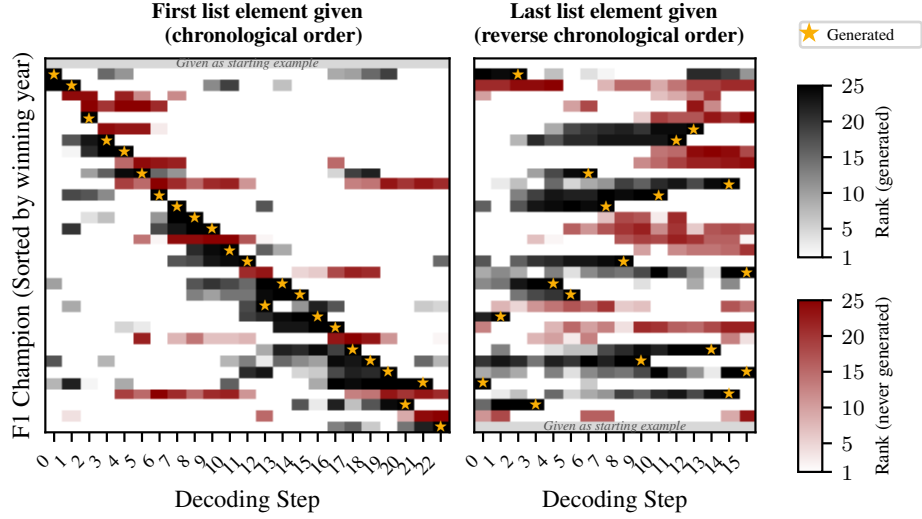}}
    \caption{Evolution of absolute ranks when generating the set of all distinct Formula 1 world champions given the first (left, \textit{Giuseppe Farina}) and most recent champion (right, \textit{Max Verstappen}) as start element.}
    \label{fig:heat_map_simple}
\end{figure*}

\subsection{Evolution of Absolute Ranks and Missing Elements}
\label{ssec:mech2}
\begin{figure}[!t]
    \centering
    \resizebox{\columnwidth}{!}{\input{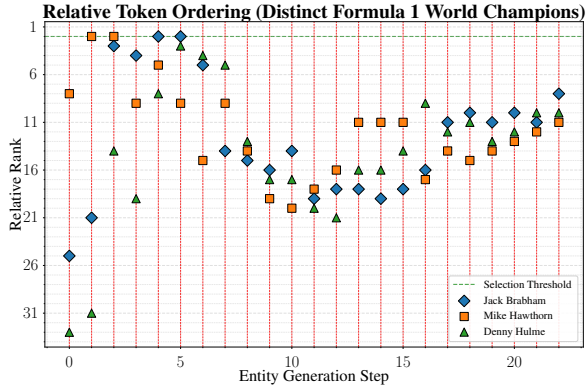}}
    \caption{Relative rank development of three different drivers that were not generated into the final output.}
    \label{fig:missing_world_champions}
\end{figure}

\textbf{In this section, we look at how entities that are ultimately missing from the generated output behave in the logit space during decoding.}
For this, we examine the much larger set of all \enquote{Formula 1 world champions} as an example and compare two runs: one with the first-ever world champion (\textit{Giuseppe Farina}) as start element and one with \textit{Max Verstappen} as start element, who is the most recent world champion w.r.t.~the knowledge cutoff dates of most LLMs.
For all champions, we track the absolute rank of their first subtoken in the sorted logit space across all forward passes in which the generation of a new entity begins.\footnote{Note that rank collisions can occur when driver names share the same first subtoken.}

Figure \ref{fig:heat_map_simple} shows the set generation process of all F1 world champions, including the absolute ranks at each entity generation start.
The rows are sorted in ascending order, starting with the earliest world champion (depending on the prefix used) in the first row.
A cell marked with a star indicates that the driver of this row was chosen at the corresponding generation step.
White cells indicate absolute positions larger than 50.
We can see that drivers that have been generated only play a minor role in subsequent decoding steps, regardless of the canonical ordering.
What we can see is that deviating from the canonical ordering by inputting \textit{Max Verstappen} as first element has the result that the model does not stick to a generation order anymore.
Furthermore, deviating from the canonical ordering leads to fewer drivers being generated by the model.

\paragraph{Rank Developments of Missing Elements}

In Figure \ref{fig:missing_world_champions}, we show the relative rank development of three different drivers that are missing from the final output when inputting the first ever champion as start element.
We can see that they ranked at high relative spots, especially in the early stages of the list generation process.
However, they were never actually selected since they never ranked top-1 at all.
This shows that LLMs can actually miss elements even though they played an actual role during decoding.
As a result, future research should focus on detecting missing elements from the logit scores to improve recall.

We can conclude three major findings when comparing the two runs.
\bracketcounter{1} Using the first champion as the start element allows the LLM to generate subsequent drivers mostly in order, as indicated by the diagonal pattern. 
\bracketcounter{2} In contrast, using the most recent champion leads to generation of drivers out of order rather than in reverse chronological order.
\bracketcounter{3} Sticking to the order leads to more drivers being successfully generated, while deviating from the canonical ordering, in this case chronological, results in more omitted elements.
This aligns well with our previous findings that LLMs are more reliable when generating following the canonical ordering of the topic.

\section{Conclusion \& Outlook}
We position our work as a starting point for further research on multi-valued relation generation and relational knowledge extraction from LLMs as KBs. 
Our mechanistic analysis suggests that multi-valued relation generation in LLMs follows three phases: \bracketcounter{1} \enquote{retrieval} of relevant entities in the form of assigning higher probabilities to tokens that are part of entities, \bracketcounter{2} internal sorting according to a relation-specific canonical ordering, and \bracketcounter{3} selection of the next element during decoding.
Ultimately, our findings and illustration highlight a general issue with the perception of LLMs: while they encode a vast amount of useful knowledge, they do not structure it in the way principled KBs are designed or in the way humans reason about this knowledge.
Instead, our experiments clearly demonstrate that autoregressive transformer architectures learn probability distributions reflecting canonical orderings seen during pretraining rather than actually incorporating the relationships inherent to multi-valued relations.
To mitigate this issue, we recommend incorporating shuffled and perturbed variants of multi-valued relations during pre-training. More broadly, our findings suggest that using LLMs as knowledge bases requires accounting for their internal ordering biases.

\section*{Limitations}
A limitation of our work arises from the fact that we rely on subtokens when identifying entities in the logit space. 
These subtokens cannot always be unambiguously mapped to their corresponding entities, especially when multiple entities share the same initial syllables or characters. 
To mitigate this issue, we primarily select entities that are distinguishable in the subtoken space for our mechanistic analyses. 
Nevertheless, some collisions may still occur, in which case we cannot reliably determine which entity the model would generate if it continued generation from that subtoken.

Furthermore, in this work, we only deal with sets that do not change too frequently.
We leave the analysis of frequently changing sets and the LLM behavior on them for future work.

\section*{Ethical Considerations}
In this work, we focus exclusively on publicly accessible and well-known topics. 
We do not incorporate any personally sensitive data that could be used to harm individuals. Furthermore, we do not aim to elicit secret or confidential knowledge that could be used in a harmful way. 
Instead, our goal is to analyze how publicly available knowledge can be elicited from LLMs.

\bibliography{anthology1,anthology2,refs}

\begin{thebibliography}{47}
\providecommand{\natexlab}[1]{#1}

\bibitem[{Abnar and Zuidema(2020)}]{abnar-zuidema-2020-quantifying}
Samira Abnar and Willem Zuidema. 2020.
\newblock \href {https://doi.org/10.18653/v1/2020.acl-main.385} {Quantifying attention flow in transformers}.
\newblock In \emph{Proceedings of the 58th Annual Meeting of the Association for Computational Linguistics}, pages 4190--4197, Online. Association for Computational Linguistics.

\bibitem[{Agarwal et~al.(2025)Agarwal, Devaguptapu, and S}]{agarwal-etal-2025-hybrid}
Ankush Agarwal, Chaitanya Devaguptapu, and Ganesh S. 2025.
\newblock \href {https://doi.org/10.18653/v1/2025.naacl-long.39} {Hybrid graphs for table-and-text based question answering using {LLM}s}.
\newblock In \emph{Proceedings of the 2025 Conference of the Nations of the Americas Chapter of the Association for Computational Linguistics: Human Language Technologies (Volume 1: Long Papers)}, pages 858--875, Albuquerque, New Mexico. Association for Computational Linguistics.

\bibitem[{AlKhamissi et~al.(2022)AlKhamissi, Li, Celikyilmaz, Diab, and Ghazvininejad}]{alkhamissi2022reviewlanguagemodelsknowledge}
Badr AlKhamissi, Millicent Li, Asli Celikyilmaz, Mona Diab, and Marjan Ghazvininejad. 2022.
\newblock \href {https://arxiv.org/abs/2204.06031} {A review on language models as knowledge bases}.
\newblock \emph{Preprint}, arXiv:2204.06031.

\bibitem[{Arslan et~al.(2024)Arslan, Ghanem, Munawar, and Cruz}]{arslan2024survey}
Muhammad Arslan, Hussam Ghanem, Saba Munawar, and Christophe Cruz. 2024.
\newblock A survey on rag with llms.
\newblock \emph{Procedia computer science}, 246:3781--3790.

\bibitem[{Berglund et~al.(2024)Berglund, Tong, Kaufmann, Balesni, Stickland, Korbak, and Evans}]{berglundreversal}
Lukas Berglund, Meg Tong, Maximilian Kaufmann, Mikita Balesni, Asa~Cooper Stickland, Tomasz Korbak, and Owain Evans. 2024.
\newblock The reversal curse: Llms trained on “a is b” fail to learn “b is a”.
\newblock In \emph{The Twelfth International Conference on Learning Representations}.

\bibitem[{Brown et~al.(2020)Brown, Mann, Ryder, Subbiah, Kaplan, Dhariwal, Neelakantan, Shyam, Sastry, Askell, Agarwal, Herbert-Voss, Krueger, Henighan, Child, Ramesh, Ziegler, Wu, Winter, Hesse, Chen, Sigler, Litwin, Gray, Chess, Clark, Berner, McCandlish, Radford, Sutskever, and Amodei}]{NEURIPS2020_1457c0d6}
Tom Brown, Benjamin Mann, Nick Ryder, Melanie Subbiah, Jared~D Kaplan, Prafulla Dhariwal, Arvind Neelakantan, Pranav Shyam, Girish Sastry, Amanda Askell, Sandhini Agarwal, Ariel Herbert-Voss, Gretchen Krueger, Tom Henighan, Rewon Child, Aditya Ramesh, Daniel Ziegler, Jeffrey Wu, Clemens Winter, and 12 others. 2020.
\newblock \href {https://proceedings.neurips.cc/paper_files/paper/2020/file/1457c0d6bfcb4967418bfb8ac142f64a-Paper.pdf} {Language models are few-shot learners}.
\newblock In \emph{Advances in Neural Information Processing Systems}, volume~33, pages 1877--1901. Curran Associates, Inc.

\bibitem[{Chen et~al.(2024)Chen, Chi, Wang, and Zhou}]{pmlr-v235-chen24i}
Xinyun Chen, Ryan~Andrew Chi, Xuezhi Wang, and Denny Zhou. 2024.
\newblock \href {https://proceedings.mlr.press/v235/chen24i.html} {Premise order matters in reasoning with large language models}.
\newblock In \emph{Proceedings of the 41st International Conference on Machine Learning}, volume 235 of \emph{Proceedings of Machine Learning Research}, pages 6596--6620. PMLR.

\bibitem[{Clark et~al.(2019)Clark, Khandelwal, Levy, and Manning}]{clark-etal-2019-bert}
Kevin Clark, Urvashi Khandelwal, Omer Levy, and Christopher~D. Manning. 2019.
\newblock \href {https://doi.org/10.18653/v1/W19-4828} {What does {BERT} look at? an analysis of {BERT}{'}s attention}.
\newblock In \emph{Proceedings of the 2019 ACL Workshop BlackboxNLP: Analyzing and Interpreting Neural Networks for NLP}, pages 276--286, Florence, Italy. Association for Computational Linguistics.

\bibitem[{Devlin et~al.(2019)Devlin, Chang, Lee, and Toutanova}]{devlin-etal-2019-bert}
Jacob Devlin, Ming-Wei Chang, Kenton Lee, and Kristina Toutanova. 2019.
\newblock \href {https://doi.org/10.18653/v1/N19-1423} {{BERT}: Pre-training of deep bidirectional transformers for language understanding}.
\newblock In \emph{Proceedings of the 2019 Conference of the North {A}merican Chapter of the Association for Computational Linguistics: Human Language Technologies, Volume 1 (Long and Short Papers)}, pages 4171--4186, Minneapolis, Minnesota. Association for Computational Linguistics.

\bibitem[{Dreger et~al.(2025)Dreger, Malek, and Eikerling}]{dreger2025large}
Max Dreger, Kourosh Malek, and Michael Eikerling. 2025.
\newblock Large language models for knowledge graph extraction from tables in materials science.
\newblock \emph{Digital Discovery}, 4(5):1221--1231.

\bibitem[{Gemma-Team et~al.(2025)}]{gemmateam2025gemma3technicalreport}
Gemma-Team et~al. 2025.
\newblock \href {https://arxiv.org/abs/2503.19786} {Gemma 3 technical report}.
\newblock \emph{Preprint}, arXiv:2503.19786.

\bibitem[{Grattafiori et~al.(2024)}]{grattafiori2024llama3herdmodels}
Aaron Grattafiori et~al. 2024.
\newblock \href {https://arxiv.org/abs/2407.21783} {The llama 3 herd of models}.
\newblock \emph{Preprint}, arXiv:2407.21783.

\bibitem[{Hu et~al.(2024)Hu, Ghosh, Nguyen, and Razniewski}]{hu2024gptkb}
Yujia Hu, Shrestha Ghosh, Tuan-Phong Nguyen, and Simon Razniewski. 2024.
\newblock Gptkb: Building very large knowledge bases from language models.
\newblock \emph{arXiv preprint arXiv:2411.04920}.

\bibitem[{Huo et~al.(2024)Huo, Yan, Hu, Yue, and Hu}]{huo-etal-2024-mmneuron}
Jiahao Huo, Yibo Yan, Boren Hu, Yutao Yue, and Xuming Hu. 2024.
\newblock \href {https://doi.org/10.18653/v1/2024.emnlp-main.387} {{MMN}euron: Discovering neuron-level domain-specific interpretation in multimodal large language model}.
\newblock In \emph{Proceedings of the 2024 Conference on Empirical Methods in Natural Language Processing}, pages 6801--6816, Miami, Florida, USA. Association for Computational Linguistics.

\bibitem[{Jain and Wallace(2019)}]{jain-wallace-2019-attention}
Sarthak Jain and Byron~C. Wallace. 2019.
\newblock \href {https://doi.org/10.18653/v1/N19-1357} {{A}ttention is not {E}xplanation}.
\newblock In \emph{Proceedings of the 2019 Conference of the North {A}merican Chapter of the Association for Computational Linguistics: Human Language Technologies, Volume 1 (Long and Short Papers)}, pages 3543--3556, Minneapolis, Minnesota. Association for Computational Linguistics.

\bibitem[{Jiang et~al.(2023)Jiang, Sablayrolles, Mensch, Bamford, Chaplot, de~las Casas, Bressand, Lengyel, Lample, Saulnier, Lavaud, Lachaux, Stock, Scao, Lavril, Wang, Lacroix, and Sayed}]{jiang2023mistral7b}
Albert~Q. Jiang, Alexandre Sablayrolles, Arthur Mensch, Chris Bamford, Devendra~Singh Chaplot, Diego de~las Casas, Florian Bressand, Gianna Lengyel, Guillaume Lample, Lucile Saulnier, Lélio~Renard Lavaud, Marie-Anne Lachaux, Pierre Stock, Teven~Le Scao, Thibaut Lavril, Thomas Wang, Timothée Lacroix, and William~El Sayed. 2023.
\newblock \href {https://arxiv.org/abs/2310.06825} {Mistral 7b}.
\newblock \emph{Preprint}, arXiv:2310.06825.

\bibitem[{JIANG et~al.(2025)JIANG, Chen, Li, Chao, LIU, and Cong}]{jiang2025fstllm}
YUE JIANG, Yile Chen, Xiucheng Li, Qin Chao, SHUAI LIU, and Gao Cong. 2025.
\newblock \href {https://openreview.net/forum?id=oyoiHf51es} {{FSTLLM}: Spatio-temporal {LLM} for few shot time series forecasting}.
\newblock In \emph{Forty-second International Conference on Machine Learning}.

\bibitem[{Jiang et~al.(2020)Jiang, Xu, Araki, and Neubig}]{jiang-etal-2020-know}
Zhengbao Jiang, Frank~F. Xu, Jun Araki, and Graham Neubig. 2020.
\newblock \href {https://doi.org/10.1162/tacl_a_00324} {How can we know what language models know?}
\newblock \emph{Transactions of the Association for Computational Linguistics}, 8:423--438.

\bibitem[{Li et~al.(2023)Li, Hui, QU, Yang, Li, Li, Wang, Qin, Geng, Huo, Zhou, Ma, Li, Chang, Huang, Cheng, and Li}]{li2023can}
Jinyang Li, Binyuan Hui, GE~QU, Jiaxi Yang, Binhua Li, Bowen Li, Bailin Wang, Bowen Qin, Ruiying Geng, Nan Huo, Xuanhe Zhou, Chenhao Ma, Guoliang Li, Kevin Chang, Fei Huang, Reynold Cheng, and Yongbin Li. 2023.
\newblock \href {https://openreview.net/forum?id=dI4wzAE6uV} {Can {LLM} already serve as a database interface? a {BI}g bench for large-scale database grounded text-to-{SQL}s}.
\newblock In \emph{Thirty-seventh Conference on Neural Information Processing Systems Datasets and Benchmarks Track}.

\bibitem[{Liu et~al.(2024{\natexlab{a}})Liu, Lin, Hewitt, Paranjape, Bevilacqua, Petroni, and Liang}]{liu-etal-2024-lost}
Nelson~F. Liu, Kevin Lin, John Hewitt, Ashwin Paranjape, Michele Bevilacqua, Fabio Petroni, and Percy Liang. 2024{\natexlab{a}}.
\newblock \href {https://doi.org/10.1162/tacl_a_00638} {Lost in the middle: How language models use long contexts}.
\newblock \emph{Transactions of the Association for Computational Linguistics}, 12:157--173.

\bibitem[{Liu et~al.(2024{\natexlab{b}})Liu, Wang, and Chen}]{liu-etal-2024-rethinking}
Tianyang Liu, Fei Wang, and Muhao Chen. 2024{\natexlab{b}}.
\newblock \href {https://doi.org/10.18653/v1/2024.naacl-long.26} {Rethinking tabular data understanding with large language models}.
\newblock In \emph{Proceedings of the 2024 Conference of the North American Chapter of the Association for Computational Linguistics: Human Language Technologies (Volume 1: Long Papers)}, pages 450--482, Mexico City, Mexico. Association for Computational Linguistics.

\bibitem[{Liu et~al.(2025)Liu, Gan, Wang, Zhang, Bo, Sun, Chen, and Zhang}]{10.1145/3696410.3714816}
Zhiqiang Liu, Chengtao Gan, Junjie Wang, Yichi Zhang, Zhongpu Bo, Mengshu Sun, Huajun Chen, and Wen Zhang. 2025.
\newblock \href {https://doi.org/10.1145/3696410.3714816} {Ontotune: Ontology-driven self-training for aligning large language models}.
\newblock In \emph{Proceedings of the ACM on Web Conference 2025}, WWW '25, page 119–133, New York, NY, USA. Association for Computing Machinery.

\bibitem[{Lu et~al.(2022)Lu, Bartolo, Moore, Riedel, and Stenetorp}]{lu-etal-2022-fantastically}
Yao Lu, Max Bartolo, Alastair Moore, Sebastian Riedel, and Pontus Stenetorp. 2022.
\newblock \href {https://doi.org/10.18653/v1/2022.acl-long.556} {Fantastically ordered prompts and where to find them: Overcoming few-shot prompt order sensitivity}.
\newblock In \emph{Proceedings of the 60th Annual Meeting of the Association for Computational Linguistics (Volume 1: Long Papers)}, pages 8086--8098, Dublin, Ireland. Association for Computational Linguistics.

\bibitem[{Mai et~al.(2024)Mai, Chu, and Paulheim}]{mai2024llms}
Huu~Tan Mai, Cuong~Xuan Chu, and Heiko Paulheim. 2024.
\newblock Do llms really adapt to domains? an ontology learning perspective.
\newblock In \emph{International Semantic Web Conference}, pages 126--143. Springer.

\bibitem[{Moeller et~al.(2025)Moeller, Tilli, Vu, and Pad{\'o}}]{moeller2025explaining}
Lucas Moeller, Pascal Tilli, Thang Vu, and Sebastian Pad{\'o}. 2025.
\newblock \href {https://openreview.net/forum?id=HUUL19U7HP} {Explaining caption-image interactions in {CLIP} models with second-order attributions}.
\newblock \emph{Transactions on Machine Learning Research}.

\bibitem[{Nagasawa et~al.(2023)Nagasawa, Heinzerling, Kokuta, and Inui}]{nagasawa-etal-2023-lms}
Haruki Nagasawa, Benjamin Heinzerling, Kazuma Kokuta, and Kentaro Inui. 2023.
\newblock \href {https://doi.org/10.18653/v1/2023.acl-srw.22} {Can {LM}s store and retrieve 1-to-n relational knowledge?}
\newblock In \emph{Proceedings of the 61st Annual Meeting of the Association for Computational Linguistics (Volume 4: Student Research Workshop)}, pages 130--138, Toronto, Canada. Association for Computational Linguistics.

\bibitem[{OpenAI et~al.(2025)}]{openai2025gptoss120bgptoss20bmodel}
OpenAI et~al. 2025.
\newblock \href {https://arxiv.org/abs/2508.10925} {gpt-oss-120b \& gpt-oss-20b model card}.
\newblock \emph{Preprint}, arXiv:2508.10925.

\bibitem[{Papaluca et~al.(2024)Papaluca, Krefl, Rodr{\'i}guez~M{\'e}ndez, Lensky, and Suominen}]{papaluca-etal-2024-zero}
Andrea Papaluca, Daniel Krefl, Sergio Rodr{\'i}guez~M{\'e}ndez, Artem Lensky, and Hanna Suominen. 2024.
\newblock \href {https://doi.org/10.18653/v1/2024.kallm-1.2} {Zero- and few-shots knowledge graph triplet extraction with large language models}.
\newblock In \emph{Proceedings of the 1st Workshop on Knowledge Graphs and Large Language Models (KaLLM 2024)}, pages 12--23, Bangkok, Thailand. Association for Computational Linguistics.

\bibitem[{Petroni et~al.(2019)Petroni, Rockt{\"a}schel, Riedel, Lewis, Bakhtin, Wu, and Miller}]{petroni-etal-2019-language}
Fabio Petroni, Tim Rockt{\"a}schel, Sebastian Riedel, Patrick Lewis, Anton Bakhtin, Yuxiang Wu, and Alexander Miller. 2019.
\newblock \href {https://doi.org/10.18653/v1/D19-1250} {Language models as knowledge bases?}
\newblock In \emph{Proceedings of the 2019 Conference on Empirical Methods in Natural Language Processing and the 9th International Joint Conference on Natural Language Processing (EMNLP-IJCNLP)}, pages 2463--2473, Hong Kong, China. Association for Computational Linguistics.

\bibitem[{Pinto et~al.(2025)Pinto, Oliveira, and Fleger}]{pinto2025exploring}
Tom{\'a}s Cerveira Da~Cruz Pinto, Hugo~Gon{\c{c}}alo Oliveira, and Chris-Bennet Fleger. 2025.
\newblock Exploring medium-sized llms for knowledge base construction.
\newblock In \emph{Proceedings of the 5th Conference on Language, Data and Knowledge}, pages 221--232.

\bibitem[{Prakash and Lee(2023)}]{prakash-lee-2023-layered}
Nirmalendu Prakash and Roy Ka-Wei Lee. 2023.
\newblock \href {https://doi.org/10.18653/v1/2023.blackboxnlp-1.22} {Layered bias: Interpreting bias in pretrained large language models}.
\newblock In \emph{Proceedings of the 6th BlackboxNLP Workshop: Analyzing and Interpreting Neural Networks for NLP}, pages 284--295, Singapore. Association for Computational Linguistics.

\bibitem[{Qin et~al.(2024)Qin, Luo, Wang, Jiang, and Sun}]{qin2024relationaldatabaseaugmentedlarge}
Zongyue Qin, Chen Luo, Zhengyang Wang, Haoming Jiang, and Yizhou Sun. 2024.
\newblock \href {https://arxiv.org/abs/2407.15071} {Relational database augmented large language model}.
\newblock \emph{Preprint}, arXiv:2407.15071.

\bibitem[{Radford et~al.(2021)Radford, Kim, Hallacy, Ramesh, Goh, Agarwal, Sastry, Askell, Mishkin, Clark, Krueger, and Sutskever}]{radford2021learningtransferablevisualmodels}
Alec Radford, Jong~Wook Kim, Chris Hallacy, Aditya Ramesh, Gabriel Goh, Sandhini Agarwal, Girish Sastry, Amanda Askell, Pamela Mishkin, Jack Clark, Gretchen Krueger, and Ilya Sutskever. 2021.
\newblock \href {https://arxiv.org/abs/2103.00020} {Learning transferable visual models from natural language supervision}.
\newblock \emph{Preprint}, arXiv:2103.00020.

\bibitem[{Ratcliff et~al.(1988)Ratcliff, Metzener et~al.}]{ratcliff1988pattern}
John~W Ratcliff, David~E Metzener, et~al. 1988.
\newblock Pattern matching: The gestalt approach.
\newblock \emph{Dr. Dobb’s Journal}, 13(7):46.

\bibitem[{Regino and dos Reis(2025)}]{regino-dos-reis-2025-llms}
Andr{\'e}~Gomes Regino and Julio~Cesar dos Reis. 2025.
\newblock \href {https://aclanthology.org/2025.genaik-1.10/} {Can {LLM}s be knowledge graph curators for validating triple insertions?}
\newblock In \emph{Proceedings of the Workshop on Generative AI and Knowledge Graphs (GenAIK)}, pages 87--99, Abu Dhabi, UAE. International Committee on Computational Linguistics.

\bibitem[{Singhania et~al.(2023)Singhania, Razniewski, and Weikum}]{singhania-etal-2023-extracting}
Sneha Singhania, Simon Razniewski, and Gerhard Weikum. 2023.
\newblock \href {https://doi.org/10.18653/v1/2023.repl4nlp-1.12} {Extracting multi-valued relations from language models}.
\newblock In \emph{Proceedings of the 8th Workshop on Representation Learning for NLP (RepL4NLP 2023)}, pages 139--154, Toronto, Canada. Association for Computational Linguistics.

\bibitem[{Suchanek et~al.(2007)Suchanek, Kasneci, and Weikum}]{yago}
Fabian~M. Suchanek, Gjergji Kasneci, and Gerhard Weikum. 2007.
\newblock \href {https://doi.org/10.1145/1242572.1242667} {Yago: a core of semantic knowledge}.
\newblock In \emph{Proceedings of the 16th International Conference on World Wide Web}, WWW '07, page 697–706, New York, NY, USA. Association for Computing Machinery.

\bibitem[{Sundararajan et~al.(2017)Sundararajan, Taly, and Yan}]{DBLP:conf/icml/SundararajanTY17}
Mukund Sundararajan, Ankur Taly, and Qiqi Yan. 2017.
\newblock \href {http://proceedings.mlr.press/v70/sundararajan17a.html} {Axiomatic attribution for deep networks}.
\newblock In \emph{Proceedings of the 34th International Conference on Machine Learning, {ICML} 2017, Sydney, NSW, Australia, 6-11 August 2017}, Proceedings of Machine Learning Research, pages 3319--3328. {PMLR}.

\bibitem[{Vrande\v{c}i\'{c} and Kr\"{o}tzsch(2014)}]{wikidata}
Denny Vrande\v{c}i\'{c} and Markus Kr\"{o}tzsch. 2014.
\newblock \href {https://doi.org/10.1145/2629489} {Wikidata: a free collaborative knowledgebase}.
\newblock \emph{Commun. ACM}, 57(10):78–85.

\bibitem[{Wahidur et~al.(2025)Wahidur, Kim, Choi, Bhatti, and Lee}]{10887211}
Rahman S.~M. Wahidur, Sumin Kim, Haeung Choi, David~S. Bhatti, and Heung-No Lee. 2025.
\newblock \href {https://doi.org/10.1109/ACCESS.2025.3542125} {Legal query rag}.
\newblock \emph{IEEE Access}, 13:36978--36994.

\bibitem[{Wang et~al.(2025)Wang, Adel, Lange, Liu, Nie, Str{\"o}tgen, and Schuetze}]{wang-etal-2025-lost-multilinguality}
Mingyang Wang, Heike Adel, Lukas Lange, Yihong Liu, Ercong Nie, Jannik Str{\"o}tgen, and Hinrich Schuetze. 2025.
\newblock \href {https://doi.org/10.18653/v1/2025.acl-long.253} {Lost in multilinguality: Dissecting cross-lingual factual inconsistency in transformer language models}.
\newblock In \emph{Proceedings of the 63rd Annual Meeting of the Association for Computational Linguistics (Volume 1: Long Papers)}, pages 5075--5094, Vienna, Austria. Association for Computational Linguistics.

\bibitem[{Wei et~al.(2020)Wei, Su, Wang, Tian, and Chang}]{wei-etal-2020-novel}
Zhepei Wei, Jianlin Su, Yue Wang, Yuan Tian, and Yi~Chang. 2020.
\newblock \href {https://doi.org/10.18653/v1/2020.acl-main.136} {A novel cascade binary tagging framework for relational triple extraction}.
\newblock In \emph{Proceedings of the 58th Annual Meeting of the Association for Computational Linguistics}, pages 1476--1488, Online. Association for Computational Linguistics.

\bibitem[{Zhang and Soh(2024)}]{zhang2024extract}
Bowen Zhang and Harold Soh. 2024.
\newblock Extract, define, canonicalize: An llm-based framework for knowledge graph construction.
\newblock In \emph{Proceedings of the 2024 Conference on Empirical Methods in Natural Language Processing}, pages 9820--9836.

\bibitem[{Zhao et~al.(2021)Zhao, Wallace, Feng, Klein, and Singh}]{zhao2021calibrateuseimprovingfewshot}
Tony~Z. Zhao, Eric Wallace, Shi Feng, Dan Klein, and Sameer Singh. 2021.
\newblock \href {https://arxiv.org/abs/2102.09690} {Calibrate before use: Improving few-shot performance of language models}.
\newblock \emph{Preprint}, arXiv:2102.09690.

\bibitem[{Zhou et~al.(2025{\natexlab{a}})Zhou, Mesgar, Friedrich, and Adel}]{zhou-etal-2025-efficient}
Wei Zhou, Mohsen Mesgar, Annemarie Friedrich, and Heike Adel. 2025{\natexlab{a}}.
\newblock \href {https://doi.org/10.18653/v1/2025.findings-naacl.54} {Efficient multi-agent collaboration with tool use for online planning in complex table question answering}.
\newblock In \emph{Findings of the Association for Computational Linguistics: NAACL 2025}, pages 945--968, Albuquerque, New Mexico. Association for Computational Linguistics.

\bibitem[{Zhou et~al.(2025{\natexlab{b}})Zhou, Li, Lu, Li, Liu, Zhang, Wang, He, Liu, and Zhang}]{zhou-etal-2025-reflection}
Yigeng Zhou, Wu~Li, Yifan Lu, Jing Li, Fangming Liu, Meishan Zhang, Yequan Wang, Daojing He, Honghai Liu, and Min Zhang. 2025{\natexlab{b}}.
\newblock \href {https://doi.org/10.18653/v1/2025.findings-acl.1221} {Reflection on knowledge graph for large language models reasoning}.
\newblock In \emph{Findings of the Association for Computational Linguistics: ACL 2025}, pages 23840--23857, Vienna, Austria. Association for Computational Linguistics.

\bibitem[{Zhu et~al.(2025)Zhu, Xie, Liu, Li, and Hu}]{zhu-etal-2025-knowledge}
Xiangrong Zhu, Yuexiang Xie, Yi~Liu, Yaliang Li, and Wei Hu. 2025.
\newblock \href {https://doi.org/10.18653/v1/2025.naacl-long.449} {Knowledge graph-guided retrieval augmented generation}.
\newblock In \emph{Proceedings of the 2025 Conference of the Nations of the Americas Chapter of the Association for Computational Linguistics: Human Language Technologies (Volume 1: Long Papers)}, pages 8912--8924, Albuquerque, New Mexico. Association for Computational Linguistics.

\end{thebibliography}

\end{document}